\documentclass[runningheads]{llncs}
\usepackage{graphicx}
\usepackage{amsmath,amssymb} 
\usepackage{color}

\usepackage[dvipsnames]{xcolor}
\usepackage{bm}
\usepackage{hyperref}
\usepackage{cleveref}
\usepackage{tikz}
\usepackage{rotating}
\usepackage{wrapfig}
\usepackage{pbox}
\usepackage{adjustbox}

\crefname{section}{sec.}{sec.}
\Crefname{section}{Sec.}{Sec.}
\crefname{figure}{fig.}{fig.}
\Crefname{figure}{Fig.}{Fig.}
\Crefname{equation}{eq.}{eq.}
\Crefname{equation}{Eq.}{Eq.}
\newcommand{\myparagraph}[1]{\vspace{0.5em}\noindent {\bf #1.}}

\newcommand{\bx}{\mathbf{x}}

\newcommand{\bp}{\mathbf{p}}

\newcommand{\methodsize}{\tiny}

\newcommand*\samethanks[1][\value{footnote}]{\footnotemark[#1]}
\begin{document}
\title{Semi-convolutional Operators for \\ Instance Segmentation} 

\titlerunning{Novotny et al.: Semi-convolutional Operators for Instance Segmentation}
%
\author{
David Novotny\thanks{Equal contribution}\inst{1,2} \and 
\mbox{Samuel Albanie}\samethanks[1]\inst{1} \and 
\mbox{Diane Larlus}\inst{2} \and
\mbox{Andrea Vedaldi}\inst{1}\\ 
}
%
\authorrunning{Novotny D., Albanie S., Larlus D., Vedaldi A.}
%

\institute{
	Visual Geometry Group, Department of Engineering Science, University of Oxford \\
	\email{\{david,albanie,vedaldi\}@robots.ox.ac.uk} \vspace{0.2cm} \and
	Computer Vision Group, NAVER LABS Europe \\
	\email{diane.larlus@naverlabs.com}
}
\maketitle              

\begin{abstract}
Object detection and instance segmentation are dominated by region-based methods such as Mask RCNN. However, there is a growing interest in reducing these problems to pixel labeling tasks, as the latter could be more efficient, could be integrated seamlessly in image-to-image network architectures as used in many other tasks, and could be more accurate for objects that are not well approximated by bounding boxes. In this paper we show theoretically and empirically that constructing dense pixel embeddings that can separate object instances cannot be easily achieved using convolutional operators. At the same time, we show that simple modifications, which we call semi-convolutional, have a much better chance of succeeding at this task. We use the latter to show a connection to Hough voting as well as to a variant of the bilateral kernel that is spatially steered by a convolutional network. We demonstrate that these operators can also be used to improve approaches such as Mask RCNN, demonstrating better segmentation of complex biological shapes and PASCAL VOC categories than achievable by Mask RCNN alone.
\end{abstract}

\section{Introduction}\label{s:intro}

\begin{figure}[t]
\includegraphics[width=\linewidth]{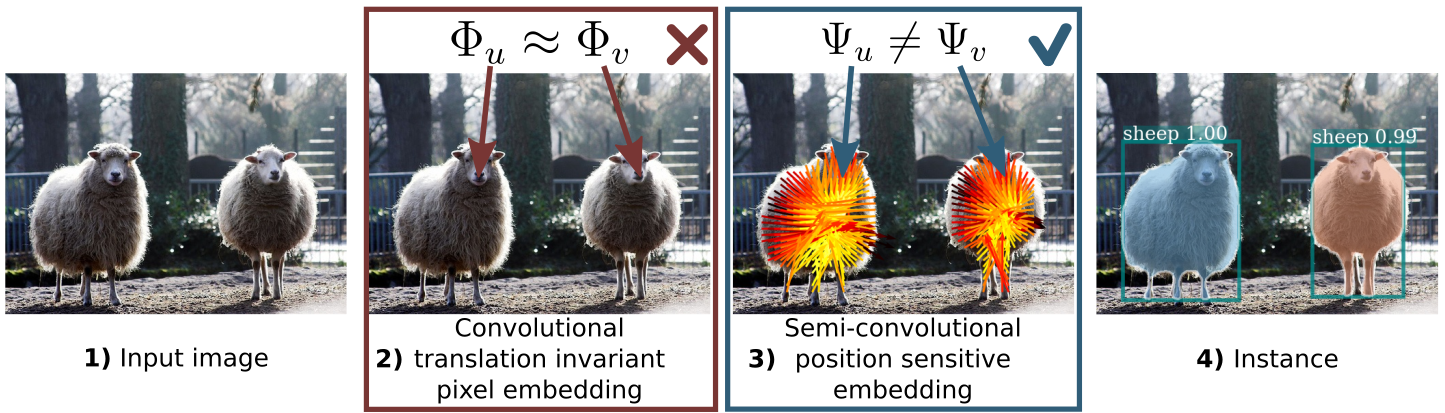}
\caption{Approaches for instance segmentation based on dense coloring via convolutional pixel embeddings cannot easily distinguishing identical copies of an object. In this paper, we propose a novel semi-convolutional embedding that is better suited for instance segmentation.\label{f:splash}}
\end{figure}

State-of-the-art methods for detecting objects in images, such as R-CNN~\cite{girshick14rcnn,girshick15fastrcnn,ren15faster}, YOLO~\cite{redmon17yolo}, and SSD~\cite{liu16ssd}, can be seen as variants of the same paradigm: a certain number of candidate image regions are proposed, either dynamically or from a fixed pool, and then a convolutional neural network (CNN) is used to decide which of these regions tightly enclose an instance of the object of interest. An important advantage of this strategy, which we call \emph{propose \& verify} (P\&V), is that it works particularly well with standard CNNs. However, P\&V also has several significant  shortcomings, starting from the fact that rectangular proposals can only approximate the actual shape of objects; segmenting objects, in particular, requires a two-step approach where, as in Mask R-CNN~\cite{he17mask}, one first detects object instances using simple shapes such as rectangles, and only then refines the detections to pixel-accurate segmentations.

An alternative to P\&V that can overcome such limitations is to label directly individual pixels with an identifier of the corresponding object occurrence. This approach, which we call \emph{instance coloring} (IC), can efficiently represent any number of objects of arbitrary shape by predicting a single label map. Thus IC is in principle much more efficient than P\&V. Another appeal of IC is that it can be formulated as an image-to-image regression problem, similar to other image understanding tasks such as denoising, depth and normal estimation, and semantic segmentation. Thus this strategy may allow to more easily build \emph{unified architectures} such as~\cite{kokkinos17uber,kendall2017multi} that can solve instance segmentations together with other problems.

Despite the theoretical benefits of IC, however, P\&V methods currently dominate in terms of overall accuracy. 
The goal of this paper is to explore some of the reasons for this gap and to suggest workarounds. 
Part of the problem may be in the nature of the dense labels. The most obvious way of coloring objects is to number them and ``paint'' them with their corresponding number. However, the latter is a global operation as it requires to be aware of all the objects in the image. CNNs, which are \emph{local and translation invariant}, may therefore be ill-suited for direct enumeration. Several authors have thus explored alternative coloring schemes  more suitable for convolutional networks. A popular approach is to assign an arbitrary color (often in the guise of a real vector) to each object occurrence, with the only requirement that different colors should be used for different objects~\cite{fathi17semantic,chandra17dense,kong18recurrent}. The resulting color \emph{affinities} can then be used to easily enumerate object a posteriori via a non-convolutional algorithm.

In this paper, we argue that even the latter technique is insufficient to make IC amenable to computation by CNNs. The reason is that, since CNNs are translation invariant, they must still assign the same color to identical copies of an object, making replicas indistinguishable by convolutional coloring. This argument, which is developed rigorously in~\cref{s:theo}, holds in the limit since in practice the receptive field size of most CNNs is nearly as large as the whole image; however, it suggests that the convolutional structure of the network is at least an unnatural fit for IC.

In order to overcome this issue, we suggest that an architecture used for IC should not be translation invariant; while this may appear to be a significant departure from convolutional networks, we also show that a small modification of standard CNNs can overcome the problem. We do so by defining \emph{semi-convolutional} operators which mix information extracted from a convolutional network with information about the global location of a pixel (\cref{s:scn,f:splash}). We train the latter (\cref{s:learn}) so that the response of the operator is similar for all pixels that belong to the same object instance, making this embedding naturally suited for IC. We show that, if the mixing function is additive, then the resulting operator bears some resemblance to Hough voting and related detection approaches. After extending the embedding to incorporate standard convolutional responses that capture appearance cues (\cref{s:traits}), we use it to induce pixel affinities and show how the latter can be interpreted as a steered version of a bilateral kernel (\cref{s:steered}). Finally, we show how such affinities can also be integrated in methods such as Mask RCNN (\cref{s:mrcnn3000}).


We assess our method with several experiments. We start by investigating the limit properties of our approach on simple synthetic data. Then, we show that our semi-convolutional feature extractor can be successfully combined with state-of-the-art approaches to tackle parsing of biological images containing overlapping and articulated organisms (\cref{s:biological}). Finally, we apply the latter to a standard instance segmentation benchmark PASCAL VOC (\cref{s:ic_pascal}). We show in all such cases that the use of semi-convolutional features can improve the performance of state-of-the-art instance segmentation methods such as Mask RCNN.

\section{Related work}\label{s:related}


The past years have seen large improvements in object detection, thanks to powerful baselines such as Faster-RCNN~\cite{ren15faster}, SSD~\cite{liu16ssd} or other similar approaches ~\cite{dai16rfcn,redmon17yolo,lin17feature}, all from the \emph{propose \& verify} strategy.

Following the success of object detection and semantic segmentation, the challenging task of instance-level segmentation has received increasing attention. Several very different families of approaches have been proposed.

\myparagraph{Proposal-based instance segmentation}
While earlier methods relied on bottom-up segmentations \cite{girshick15fastrcnn,dai15convolutional}, the vast majority of recent instance-level approaches combine segment proposals together with powerful object classifiers.
In general, they implement a multi-stage pipeline that first generates region proposals or class agnostic boxes, and then classifies them \cite{ladicky10what,hariharan14simultaneous,chen15multi,pinheiro15learning,dai16instance-aware,pinheiro16learning,liang16reversible}.
For instance DeepMask~\cite{pinheiro15learning} and follow-up approaches~\cite{pinheiro16learning,dai16instance-sensitive} learn to propose segment candidates that are then classified.
The MNC approach \cite{dai16instance-aware}, based on Faster-RCNN \cite{ren15faster}, repeats this process twice \cite{dai16instance-aware} while \cite{liang16reversible} does it multiple times. \cite{hayder17boundary} extends \cite{dai16instance-aware} to model the shape of objects.
The fully convolutional instance segmentation method of \cite{li17fully} also combines segmentation proposal and object detection using a position sensitive score map.

Some methods start with semantic segmentation first, and then cut the regions obtained for each category into multiple instances \cite{kirillov17instance,bai17deep,liu17sgn}, possibly involving higher-order CRFs \cite{arnab17pixelwise}.

Among the most successful methods to date, Mask-RCNN~\cite{he17mask} extends Faster R-CNN~\cite{ren15faster} with a small fully convolutional network branch~\cite{long15fcn} producing segmentation masks for each region of interest predicted by the detection branch. Despite its outstanding results, Mask-RCNN does not come without shortcomings: it relies on a small and predefined set of region proposals and non-maximum suppression, making it less robust to strong occlusions, crowded scenes, or objects with fundamentally non-rectangular shapes (see detailed discussion in \cref{s:theo}).

\myparagraph{Instance-sensitive embeddings}
Some works have explored the use of pixel-level embeddings in the context of clustering tasks, employing them as a soft, differentiable proxy for cluster assignments~\cite{wang15deep,harley16learning,fathi17semantic,brabandere17semantic,newell17associative,kong18recurrent}. This is reminiscent of unsupervised image segmentation approaches~\cite{shi00normalized,felzenszwalb04efficient}. It has been used for body joints~\cite{newell17associative}, semantic segmentation~\cite{harley17segmentation,harley16learning,chandra17dense} and optical flow~\cite{harley17segmentation}, and, more relevant to our work, to instance segmentation~\cite{fathi17semantic,brabandere17semantic,chandra17dense,kong18recurrent}.  

The goal of this type of approaches is to bring points that belong to the same instance close to each other in an embedding space, so that the decision for two pixels to belong to the same instance can be directly measured by a simple distance function. Such an embedding requires a high degree of invariance to the interior appearance of objects. 

Among the most recent methods, \cite{fathi17semantic} combines the embedding with a greedy mechanism to select seed pixels, that are used as starting points to construct instance segments. 
\cite{chandra17dense} connects embeddings, low rank matrices and densely connected random fields.
\cite{kong18recurrent} embeds the pixels and then groups them into instances with a variant of mean-shift that is implemented as a recurrent neural network.
All these approaches are based on convolutions, that are local and translation invariant by construction, and consequently are inherently ill-suited to distinguish several identical instances of the same object (see more details about the convolutional coloring dilemma in \cref{s:theo}). A recent work \cite{kendall2017multi} employs position sensitive convolutional embeddings that regress the location of the centroid of each pixel's instance. We mainly differ by allowing embeddings to regress an unconstrained representative point of each instance.

Among other approaches using a clustering component, \cite{silberman14instance} leverages a coverage loss and \cite{zhang15monocular,tighe14scene,uhrig16pixel} make use of depth information.
In particular, \cite{uhrig16pixel} trains a network to predict each pixel direction towards its instance center along with monocular depth and semantic labeling. Then template matching and proposal fusion techniques are applied.

\myparagraph{Other instance segmentation approaches}
Several methods \cite{pinheiro15learning,pinheiro16learning,liang15proposal,hu17fastmask} move away from box proposals and use Faster-RCNN~\cite{ren15faster} to produce ``centerness'' scores on each pixel instead. They directly predict the mask of each object in a second stage. An issue with such approaches is that objects do not necessarily fit in the receptive fields. 

Recurrent approaches sequentially generate a list of individual segments. For instance, \cite{andriluka16end} uses an LSTM for detection with a permutation invariant loss while \cite{romera16recurrent} uses an LSTM to produce binary segmentation masks for each instance. \cite{ren17end} extends \cite{romera16recurrent} by refining segmentations in each window using a box network.
These approaches are slow and do not scale to large and crowded images.

Some approaches use watershed algorithms. \cite{bai17deep} predicts pixel-level energy values and then partition the image with a watershed algorithm. \cite{kirillov17instance} combines a watershed algorithm with an instance aware boundary map. Such methods create disconnected regions, especially in the presence of occlusion.

\section{Method}\label{s:method}

\subsection{Semi-convolutional networks for instance coloring}\label{s:scn}

Let $\bx\in\mathcal{X}=\mathbb{R}^{H\times W\times 3}$ be an image and $u \in \Omega = \{1,\dots,H\}\times\{1,\dots,W\}$ a pixel. In instance segmentation, the goal is to map the image to a collection $\mathcal{S}_\bx=\{S_1,\dots,S_{K_\bx}\} \subset 2^\Omega$ of image regions, each representing an occurrence of an object of interest. The symbol $S_0 = \Omega - \cup_k S_k$ will denote the complementary region, representing background. The regions as well as their number are a function of the image and the goal is to predict both.

In this paper, we are interested in methods that reduce instance segmentation to a pixel-labeling problem. Namely, we seek to learn a function $\Phi :\mathcal{X} \rightarrow \mathcal{L}^\Omega$ that associates to each pixel $u$ a certain label $\Phi_u(\bx) \in \mathcal{L}$ so that, as a whole, labels encode the segmentation $\mathcal{S}_\bx$. Intuitively, this can be done by painting different regions with different ``colors'' (aka pixel labels) making objects easy to recover in post-processing. We call this process \emph{instance coloring} (IC).

A popular IC approach is to use real vectors $\mathcal{L}=\mathbb{R}^d$ as colors, and then require that the colors of different regions are sufficiently well separated. Formally, there should be a margin $M > 0$ such that:
\begin{equation}\label{e:margin}
 \forall u,v \in \Omega:
 \quad
 \begin{cases}
  \|\Phi_u(\bx) - \Phi_v(\bx)\| \leq 1 - M, & \exists k : u,v\in S_k,\\
  \|\Phi_u(\bx) - \Phi_v(\bx)\| \geq 1 + M, & \text{otherwise}.
 \end{cases}
\end{equation}
If this is the case, clustering colors trivially reconstructs the regions.

Unfortunately, it is difficult for a convolutional operator $\Phi$ to satisfy constraint~\eqref{e:margin} or analogous ones. While this is demonstrated formally in~\cref{s:theo}, for now an intuition suffices: if the image contains replicas of the same object, then a convolutional network, which is translation invariant, must assign the same color to each copy.

\begin{figure}[t]
\begin{center}
\includegraphics[height=2cm]{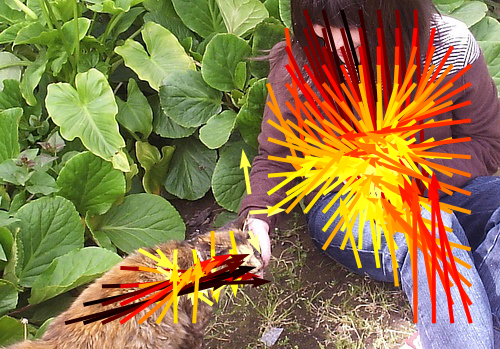}
\includegraphics[height=2cm]{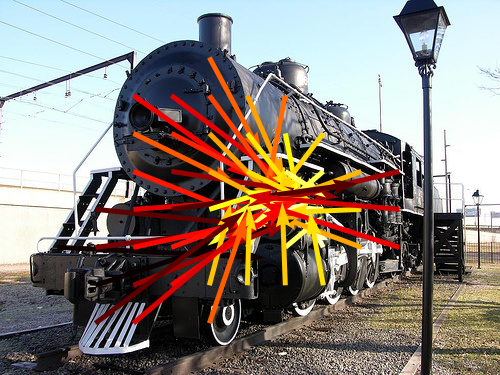}
\includegraphics[height=2cm]{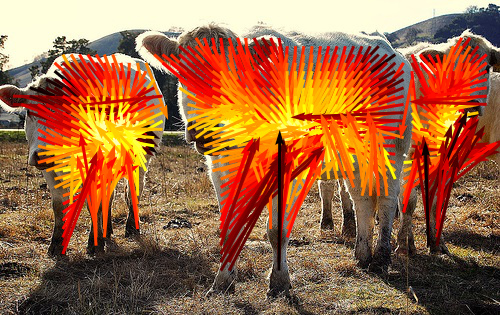}
\includegraphics[height=2cm]{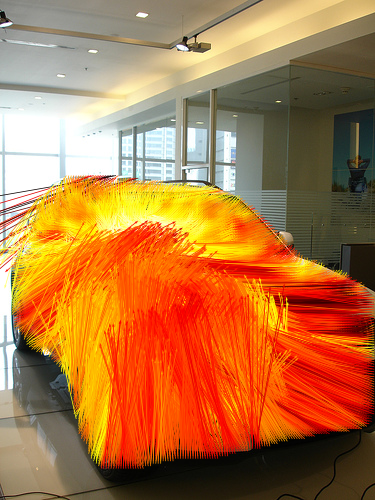}
\includegraphics[height=2cm]{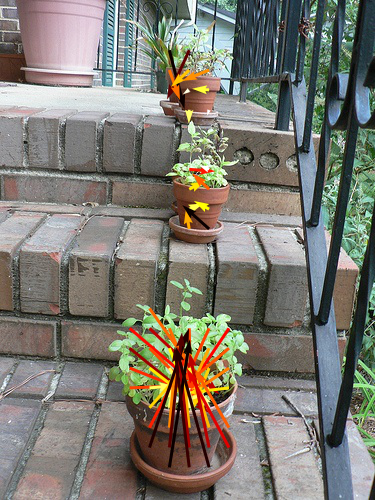}
\\
\includegraphics[height=2.06cm]{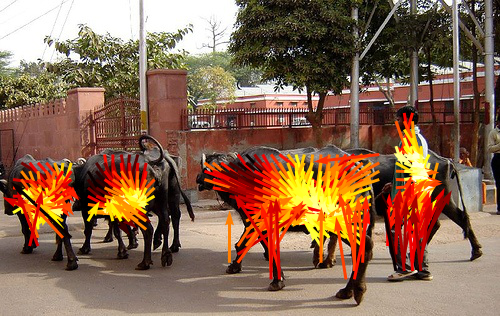}
\includegraphics[height=2.06cm]{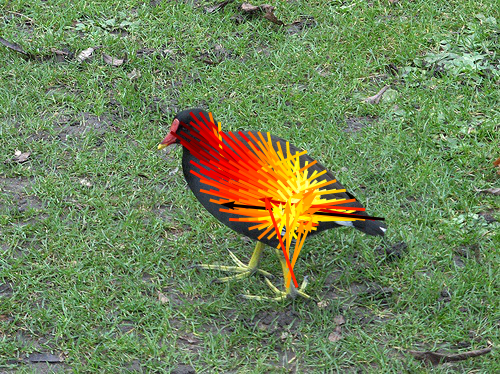}
\includegraphics[height=2.06cm]{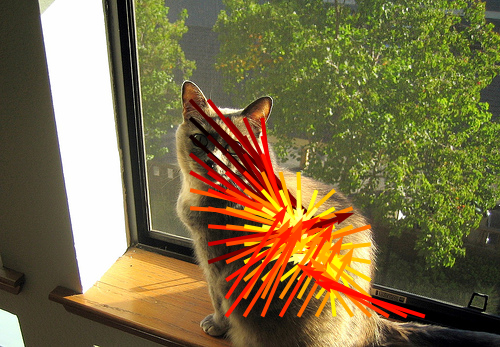}
\includegraphics[height=2.06cm]{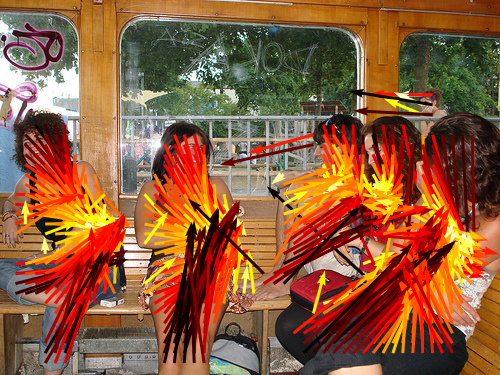}
\end{center}
\caption{\textbf{Semi-convolutional embedding.}
The first two dimensions of the embedding $\Phi_u(\bx)$ are visualized as arrows
starting from the corresponding pixel location $u$. Arrows from the
same instance tend to point towards a \textit{instance-specific} location $c_k$. \label{f:arrows}}
\end{figure}

If convolutional operators are inappropriate, then, we must abandon them in favor of non-convolutional ones. While this sounds complex, we suggest that very simple modifications of convolutional operators, which we call \emph{semi-convolutional}, may suffice. In particular, if $\Phi_u(\bx)$ is the output of a convolutional operator at pixel $u$, then we can construct a non-convolutional response by mixing it with information about the pixel location. Mathematically, we can define a semi-convolutional operator as:
\begin{equation}\label{e:semi}
  \Psi_u(\bx) = f(\Phi_u(\bx),u)
\end{equation}
where $f : \mathcal{L} \times \Omega \rightarrow \mathcal{L}'$ is a suitable mixing function. As our main example of such an operator, we consider a particularly simple type of mixing function, namely addition. With it, \cref{e:semi} specializes to:
\begin{equation}\label{e:semi2}
  \Psi_u(\bx) = \Phi_u(\bx) + u, \qquad \Phi_u(\bx) \in \mathcal{L} = \mathbb{R}^2.
\end{equation}
While this choice is restrictive, it has the benefit of having a very simple interpretation. Suppose in fact that the resulting embedding can perfectly separate instances, in the sense that $\Psi_u(\bx) = \Psi_v(\bx) \Leftrightarrow \exists k : (u,v)\in S_k$. Then for all the pixels of the region $S_k$ we can write in particular:
\begin{equation}\label{e:arrows}
 \forall u \in S_k : \quad \Phi_u(\bx) + u = c_k
\end{equation}
where $c_k\in \mathbb{R}^2$ is an \emph{instance-specific point}. In other words, we see that the effect of learning this semi-convolutional embedding for instance segmentation is to predict a \emph{displacement field} $\Phi(\bx)$ that maps all pixels of an object instance to an instance-specific centroid $c_k$. An illustration of the displacement field can be found \cref{f:arrows}.

\subsubsection{Relation to Hough voting and implicit shape models.} \Cref{e:semi2,e:arrows} are reminiscent of well known detection methods in computer vision: Hough voting~\cite{duda72use,ballard87generalizing} and implicit shape model (ISM)~\cite{leibe03interleaved}. Recall that both of these methods map image patches to votes for the parameters $\theta$ of possible object occurrences. In simple cases, $\theta\in\mathbb{R}^2$ can be the centroid of an object, and casting votes may have a form similar to~\cref{e:arrows}.

This establishes, a clear link between voting-based methods for object detection and coloring methods for instance segmentation. At the same time, there are significant differences. First, the goal here is to group pixels, not to reconstruct the parameters of an object instance (such as its centroid and scale). \Cref{e:semi2} may have this interpretation, but the more general version~\cref{e:semi} does not. Second, in methods such as Hough or ISM the centroid is defined a-priori as the actual center of the object; here the centroid $c_k$ has no explicit meaning, but is automatically inferred as a useful but arbitrary reference point. Third, in traditional voting schemes voting integrates local information extracted from individual patches; here the receptive field size of $\Phi_u(\bx)$ may be enough to comprise the whole object, or more. The goal of \cref{e:semi,e:semi2} is not to pool local information, but to solve a representational issue.

\subsection{Learning additive semi-convolutional features}\label{s:learn}

Learning the semi-convolutional features of~\cref{e:semi} can be formulated in many different ways. Here we adopt a simple direct formulation inspired by~\cite{brabandere17semantic} and build a loss by considering, for each image $\bx$ and instance $S\in\mathcal{S}$ in its segmentation, the distance between the embedding of each pixel $u\in S$ and the segment-wise mean of these embeddings:
\begin{equation}\label{e:loss1}
\mathcal{L}(\Psi|\bx,\mathcal{S})
=
\sum_{S\in\mathcal{S}}
\frac{1}{|S|}
\sum_{u \in S}
\left\|
 \Psi_u(\bx)-
 \frac{1}{|S|} \sum_{u \in S} \Psi_u(\bx)
\right\|.
\end{equation}
Note that while this quantity resembles the variance of the embedding values for each segment, it is not as the distance is not squared; this was found to be more robust. 

Note also that this loss is simpler than the margin condition~\eqref{e:margin} and than the losses proposed in~\cite{brabandere17semantic}, which resemble~\eqref{e:margin} more closely. In particular, this loss only includes an ``attractive'' force which encourages embeddings for each segment to be all equal to a certain mean value, but does not explicitly encourage different segments to be assigned different embedding values. While this can be done too, empirically we found that minimizing \cref{e:loss1} is sufficient to learn good additive semi-convolutional embeddings.

\subsection{Coloring instances using individuals' traits}\label{s:traits}

In practice, very rarely an image contains exact replicas of a certain object. Instead, it is more typical for different occurrences to have some distinctive individual traits. For example, different people are generally dressed in different ways, including wearing different colors. In instance segmentation, one can use such cues to tell right away an instance from another. Furthermore, these cues can be extracted by conventional convolutional operators.

In order to incorporate such cues in our additive semi-convolutional formulation, we still consider the expression $\Psi_u(x) = \hat u + \Phi_u(\bx)$. However, we relax $ \Phi_u(\bx)\in\mathbb{R}^d$ to have more than two dimensions $d > 2$. Furthermore, we define $\hat u$ as the pixel coordinates of $u$, $u_x$ and $u_y$, extended by zero padding:
\begin{equation}\label{e:uapp}
\hat u = \begin{bmatrix} u_x & u_y & 0 & \hdots & 0 \end{bmatrix}^\top \in \mathbb{R}^d.
\end{equation}
In this manner, the last $d-2$ dimensions of the embedding work as conventional convolutional features and can extract instance-specific traits normally.

\subsection{Steered bilateral kernels}\label{s:steered}

The pixel embedding vectors $\Psi_u(\bx)$ must ultimately be decoded as a set of image regions. Again, there are several possible strategies, starting from simple $K$-means clustering, that can be used to do so. In this section, we consider transforming embeddings in an affinity matrix between two pixels, as the latter can be used in numerous algorithms.

In order to define the affinity between pixels $u,v\in\Omega$, consider first the Gaussian kernel
\begin{equation}\label{e:gauss}
K(u,v) = \exp\left( - \frac{\|\Psi_u(\bx) - \Psi_{v}(\bx)\|^2}{2}\right).
\end{equation}
If the augmented embedding~\cref{e:uapp} is used in the definition of $\Psi_u(\bx) = \hat u + \Phi_u(\bx)$, we can split $\Phi_u(\bx)$ into a geometric part $\Phi^g_u(\bx) \in \mathbb{R}^2$ and an appearance part $\Phi^a_u(\bx)\in\mathbb{R}^{d-2}$ and expand this kernel as follows:
\begin{equation}\label{e:gauss2}
K(u,v) =
\exp
\left(
{- \frac{\|(u + \Phi^g_u(\bx)) - (v + \Phi^g_{v}(\bx))\|^2}{2}}
\right)
\exp
\left(
{- \frac{\|\Phi_u^a(\bx) - \Phi_{v}^a(\bx)\|^2}{2}}
\right).
\end{equation}
It is interesting to compare this definition to the one of the \emph{bilateral kernel}:\footnote{In the bilateral kernel, a common choice is to set $\Phi^a_u(\bx) = \bx_u \in\mathbb{R}^3$ as the RGB triplet for the appearance features.}
\begin{equation}\label{e:bil}
K_\text{bil}(u,v) =
\exp
\left({- \frac{\|u - v\|^2}{2}}
\right)
\exp
\left(
{- \frac{\|\Phi_u^a(\bx) - \Phi_{v}^a(\bx)\|^2}{2}}
\right).
\end{equation}
The bilateral kernel is very popular in many applications, including image filtering and mean shift clustering. The idea of the bilateral kernel is to consider pixels to be similar if they are close in both space and appearance. Here we have shown that kernel~\eqref{e:gauss2} and hence kernel~\eqref{e:gauss} can be interpreted as a generalization of this kernel where spatial locations are steered (distorted) by the network to move pixels that belong to the same underlying object instance closer together.

In a practical implementation of these kernels, vectors should be rescaled before being compared, for example in order to balance spatial and appearance components. In our case, since embeddings are trained end-to-end, the network can learn to perform this balancing automatically, but for the fact that~\eqref{e:arrows} implicitly defines the scaling of the spatial component of the kernel. Hence, we modify~\cref{e:gauss} in two ways: by introducing a learnable scalar parameter $\sigma$ and by considering a Laplacian rather than a Gaussian kernel:
\begin{equation}\label{e:aff}
K_\sigma(u,v) = \exp\left( - \frac{\|\Psi_u(\bx) - \Psi_{v}(\bx)\|}{\sigma}\right).
\end{equation}
This kernel is more robust to outliers (as it uses the Euclidean distance rather than its square) and is still positive definite~\cite{feragen15geodesic}. In the next section we show an example of how this kernel can be used to perform instance coloring.  

\subsection{Semi-convolutional Mask-RCNN}\label{mrcnnplus}

The semi-convolutional framework we proposed in \cref{s:scn} is very generic and can be combined with many existing approaches.
Here, we describe how it can be combined with the Mask-RCNN (MRCNN) framework~\cite{he17mask}, the current state-of-the-art in instance segmentation.

MRCNN is based on the RCNN \textit{propose \& verify} strategy and first produces a set of rectangular regions $\mathcal{R}$, where each rectangle $R \in \mathcal{R}$ tightly encloses an instance candidate.
Then a fully convolutional network (FCN) produces foreground/background segmentation inside each region candidate. In practice, it labels every pixel $u_i$ in $R$ with a foreground score logit $s(u_i)\in\mathbb{R}$.
However, this is not an optimal strategy for articulated objects or occluded scenes (as validated in \cref{s:biological}), as it is difficult for a standard FCN to perform individual foreground/background predictions. Hence we leverage our pixel-level translation sensitive embeddings in order to improve the quality of the  predictions $s(u_i)$.

\myparagraph{Extending MRCNN} \label{s:mrcnn3000}  
Our approach is based on two intuitions: first, some points are easier to be recognized as foreground than others, and, second, once one such \emph{seed point} has been determined, its affinity with other pixels can be used to cut out the foreground region.

In practice, we first identify a seed pixel $u_s$ in each region $R$ using the MRCNN foreground confidence score map $\mathbf{s} = [ s(u_1), \dots, s(u_{|R|})]$. We select the \emph{most confident} seed point as $u_s = \operatorname{argmax}_{1\leq i \leq |R|} s(u_i)$, evaluate the steered bilateral kernel $K_\sigma(u_s,u)$ after extracting the embeddings $\Psi_{u_s}$ for the seed and $\Psi_{u_i}$ of each pixel $u_i$ in the region, and then defining updated scores $\hat s(u_i)$ as  
$\hat s(u_i) = s(u_i) + \log{K_\sigma(u_s,u_i)}$. The combination of the scores and the kernel is performed in the log-space
due to improved numerical stability.
The final per-pixel foreground probabilities are obtained as in \cite{he17mask} with $\text{sigmoid}(\hat s(u_i))$.

The entire architecture \textemdash the region selection mechanism, the foreground prediction, and the pixel-level embedding \textemdash is trained end-to-end. For differentiability, this requires the following modifications: 
we replace the maximum operator with a soft maximum over the scores $\bp_s = \text{softmax}(\mathbf{s})$ and we obtain the seed embedding $\Psi_{u_s}$ as the expectation over the embeddings $\Psi_u$
under the probability density $\bp_s$.
The network optimizer minimizes, together with the MRCNN losses, the image-level embedding loss $\mathcal{L}(\Psi|\bx,\mathcal{S})$ 
and further attaches a secondary binary cross entropy loss that, similar to the MRCNN mask predictor,
minimizes binary cross entropy between the kernel output $K_\sigma(u_s,u_i)$ and the ground truth instance masks.

The predictors of our semi-convolutional features $\Psi_u$ were implemented as an output of a shallow subnetwork, shared between all the FPN layers. This subnet consists of a 256-channel 1$\times$1 convolutional filter followed by ReLU and a final 3$\times$3 convolutional filter producing $D=8$ dimensional embedding $\Psi_u$. 
Due to an excessive sensitivity of the RPN component to perturbations of the underlying FPN representation, we downscale the gradients that are generated by the shallow subnetwork and received by the shared FPN tensors by a factor of 10.

\subsection{The convolutional coloring dilemma}\label{s:theo}

In this section, we prove some properties of convolutional operators in relation to solving instance segmentation problems. In order to do this, we need to start by formalizing the problem.

We consider signals (images) of the type $\bx : \Omega \rightarrow \mathbb{R}$, where the domain $\Omega$ is either $\mathbb{Z}^m$ or $\mathbb{R}^m$.\footnote{We assume that the domain extends to infinity to avoid having to deal explicitly with boundary conditions.} In segmentation, we are given a family $\bx\in\mathcal{X}$ of such signals, each of which is associated to a certain partition $\mathcal{S}_\bx = \{S_1,\dots,S_{K_\bx}\}$ of the domain $\Omega$. The goal is to construct a \emph{segmentation algorithm}  $\mathcal{A} : \bx \mapsto \mathcal{S}_\bx$ that computes this function. We look in particular at algorithms that pre-process the signal by assigning a label $\Phi_u(\bx)\in\mathcal{L}$ to each point $u\in\Omega$ of the domain. Furthermore, we assume that this labeling operator $\Phi$ is \emph{local and translation invariant}\footnote{We say that $\Phi$ is translation invariant if $\Phi_u(\bx(\cdot - \tau))= \Phi_{u-\tau}(\bx)$ for all translations $\tau \in \Omega$.
We say that it is also local if there exists a constant $M > 0$ such that $x_u=x_{u'}$ for all $|u-u'|<M$ implies that $\Phi_u(\bx) = \Phi_u'(\bx)$.} so as to be implementable with a convolutional neural network.

There are two families of algorithms that can be used to segment signals in this manner, discussed next.

\myparagraph{Propose \& verify} The first family of algorithms submits all possible regions $S_r \subset \Omega$, indexed for convenience by a variable $r$, to a labeling function $\Phi_r(\bx) \in \{0,1\}$ that \emph{verifies} which ones belong to the segmentation $\mathcal{S}_\bx$ (i.e.\ $\Phi_r(\bx) =1 \Leftrightarrow S_r \in \mathcal{S}_\bx$). Since in practice it is not possible to test all possible subsets of $\Omega$, such an algorithm must focus on a smaller set of proposal regions. A typical choice is to consider all translated squares (or rectangles) $S_u = [-H,H]^m + u$. Since the index variable $u\in\Omega$ is now a translation, the operator $\Phi_u(\bx)$ has the form discussed above, although it is not necessarily local or translation invariant.

\myparagraph{Instance coloring} The second family of approaches directly colors (labels) pixels with the index of the corresponding region, i.e.\ %
$
\Phi_u(\bx) = k \Leftrightarrow u \in S_k.
$
Differently from P\&V, this can efficiently represent arbitrary shapes. However, the map $\Phi$ needs implicitly to decide which number to assign to each region, which is a global operation. Several authors have sought to make this more amenable to convolutional networks. 
A popular approach~\cite{fathi17semantic,brabandere17semantic} is to color pixels arbitrarily (for example using vector embeddings) so that similar colors are assigned to pixels in the same region and different colors are used between regions, as already detailed in~\cref{e:margin}.

\setlength{\intextsep}{0pt}
\setlength{\columnsep}{1em}
\begin{wrapfigure}{r}{0pt}
\begin{tikzpicture}[scale=0.5]
  \foreach \x in {-2,-1,0,1,2} { \draw (2*\x-1,0) -- (2*\x,1); \draw (2*\x,1) -- (2*\x+1,0); };
    \foreach \x in {-2,-1,0,1,2} {
      \draw [thick,arrows={to-to}] (2*\x-1,-.3) -- node[below] {$S_{\x}$}(2*\x+1,-.3) ;};
  \draw [thick,-latex](-5,0) -- (5,0) node[right]{$u$};
  \draw [thick,-latex](0,0) -- (0,1.4);
\end{tikzpicture}
\end{wrapfigure}

\myparagraph{Convolutional coloring dilemma} Here we show that, even with the variants discussed above, IC cannot be approached with convolutional operators even for cases where these would work with P\&V.
 
We do so by considering a simple 1D example. Let $\bx$ be a signal of period 2 (i.e. $x_{u+2} = x_u$) where for $u\in[-1,1]$ the signal is given by $x_{u} = \min(1-u,1+u)$. Suppose that the segmentation associated to $\bx$ is $\mathcal{S}=\{[-1,1]+2k,k\in\mathbb{Z}\}$. If we assume that a necessary condition for a coloring-based algorithm is that at least some of the regions are assigned different colors, we see that this cannot be achieved by a convolutional operator. In fact, due to the periodicity of $\bx$, any translation invariant function will assign exactly the same color to pixels $2k,k\in\mathbb{Z}$. Thus \emph{all} regions have at least one point with the same color.

On the other hand, this problem can be solved by P\&V using the proposal set $\{[-1,1]+u,u\in\Omega\}$  and the local and translation invariant verification function $\Phi_u(\bx) = [x_u = 1]$, which detects the center of each region.

The latter is an extreme example of a convolutional coloring dilemma: namely, a local and translation invariant operator will naturally assign the same color to identical copies of an object even if when they are distinct occurrences (c.f. interesting concurrent work that explores related convolutional dilemmas \cite{liu2018intriguing}).

\myparagraph{Solving the dilemma} \label{s:solving} Solving the coloring dilemma can be achieved by using operators that are \emph{not} translation invariant. In the counterexample above, this can be done by using the semi-convolutional function $\Phi_u(x) = u + (1 - x_u) \dot x_u$. It is easy to show that $\Phi_u(x)=2k$ colors each pixel $u \in S_k = [-1,1]+2k$ with twice the index of the corresponding region by moving each point $u$ to the center of the closest region. This works because such displacements can be computed by looking only locally, based on the shape of the signal.

\begin{figure}[t]
\begin{center}
\pbox[t]{0.24\textwidth}{
\includegraphics[width=0.25\textwidth,clip,trim=0cm 0 0cm 0cm]{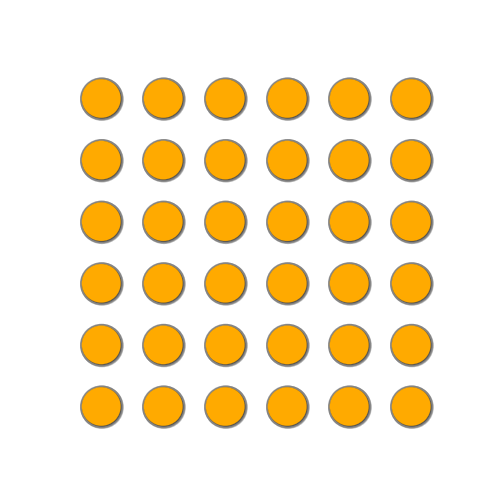}\\
\centering (a) Training image
}
\pbox[t]{0.24\textwidth}{
\includegraphics[width=0.25\textwidth,clip,trim=0cm 0 0cm 0cm]{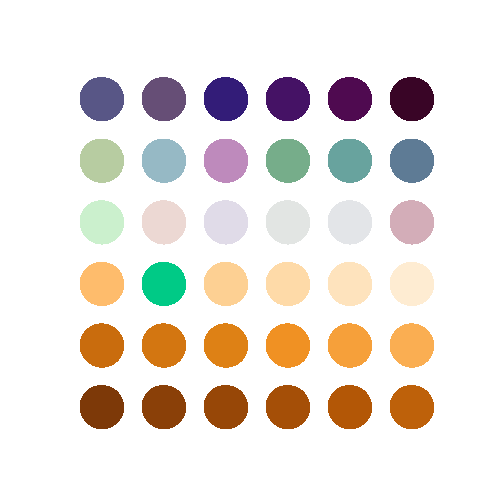}\\
\centering (b) GT instance labels
}
\pbox[t]{0.24\textwidth}{
\includegraphics[width=0.25\textwidth,clip,trim=0cm 0 0cm 0cm]{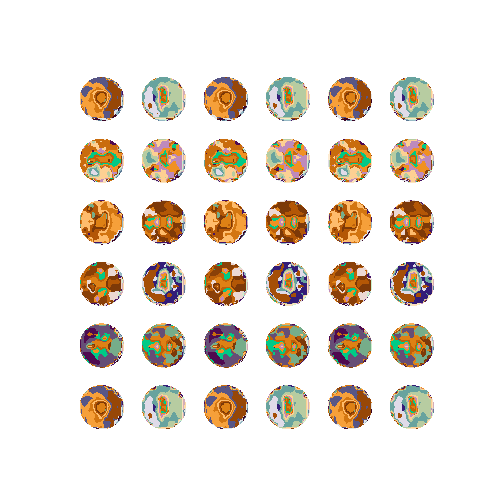}\\
\centering (c) Convolutional embedding
}
\pbox[t]{0.24\textwidth}{
\includegraphics[width=0.25\textwidth,clip,trim=0cm 0 0cm 0cm]{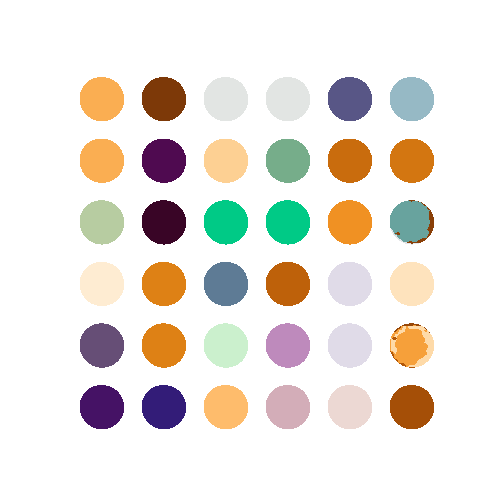}\\
\centering (d) Semi-conv. embedding (ours)
}
\end{center}
\caption{\textbf{Experiment on synthetic data.} An instance segmentation pixel embedding is trained for a synthetic training image consisting of a regular dot pattern (a). After training a model on that image, the produced embeddings are clustered using $k$-means, encoding the corresponding cluster assignments with consistent pixel colors.
A standard convolutional embedding (c) cannot successfully embed each dot into a unique location due to its translational invariance. Our proposed semi-convolutional operator (d) naturally embeds dots with identical appearance but distinct location into distinct regions in the feature space and hence allows for successful clustering of the instances. }\label{f:elegans}
\label{f:dots}
\end{figure}

\section{Experiments}\label{s:experiments}

We first conduct experiments on synthetic data in order to clearly demonstrate inherent limitations of convolutional operators for the task of instance segmentation. In the ensuing parts we demonstrate benefits of the semi-convolutional operators on a challenging scenario with a high number of  overlapping articulated instances and finally we compare to the competition on a standard instance segmentation benchmark.

\begin{figure}[t]
\begin{center}
\includegraphics[width=0.12\textwidth,clip,trim=3cm 0 3cm 3cm]{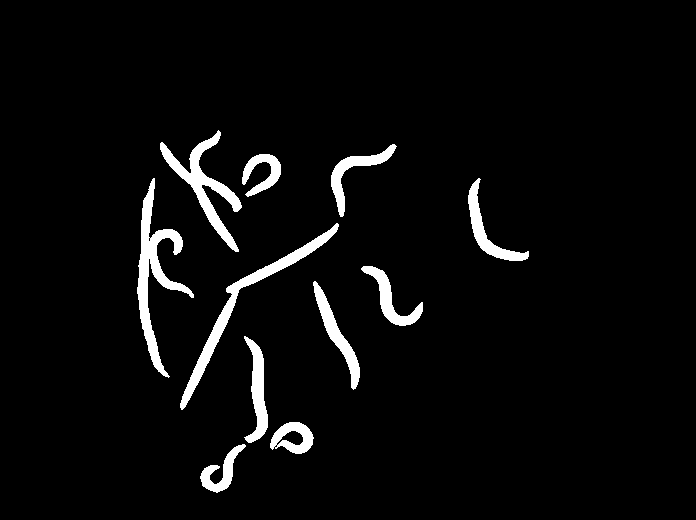}%
\includegraphics[width=0.12\textwidth,clip,trim=3cm 0 3cm 3cm]{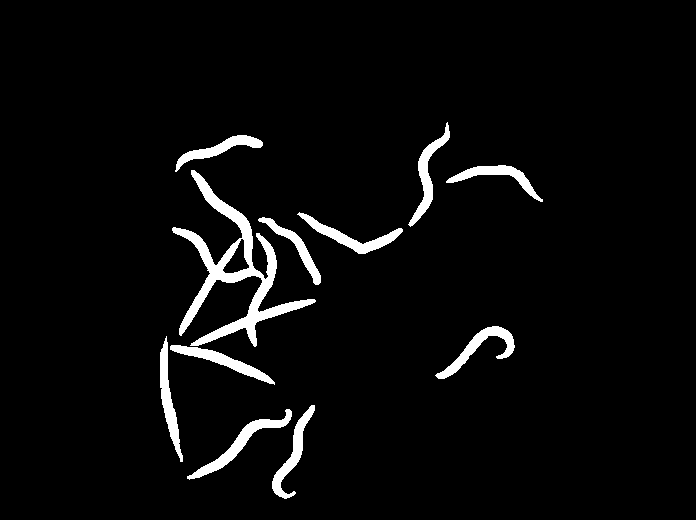}%
\includegraphics[width=0.12\textwidth,clip,trim=3cm 0 3cm 3cm]{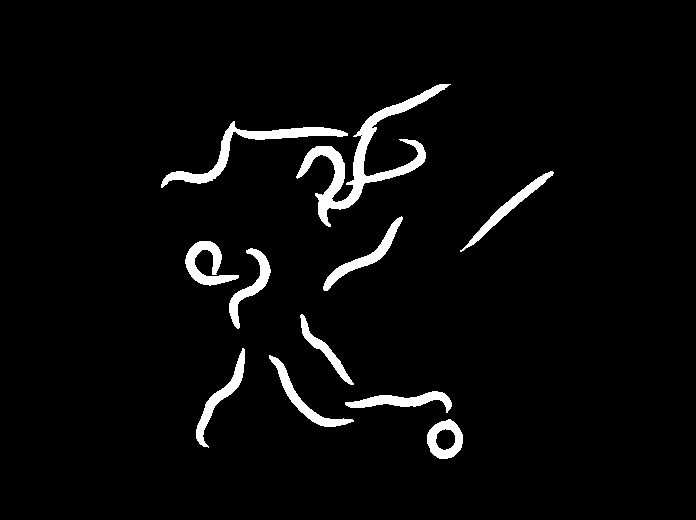}%
\includegraphics[width=0.12\textwidth,clip,trim=3cm 0 3cm 3cm]{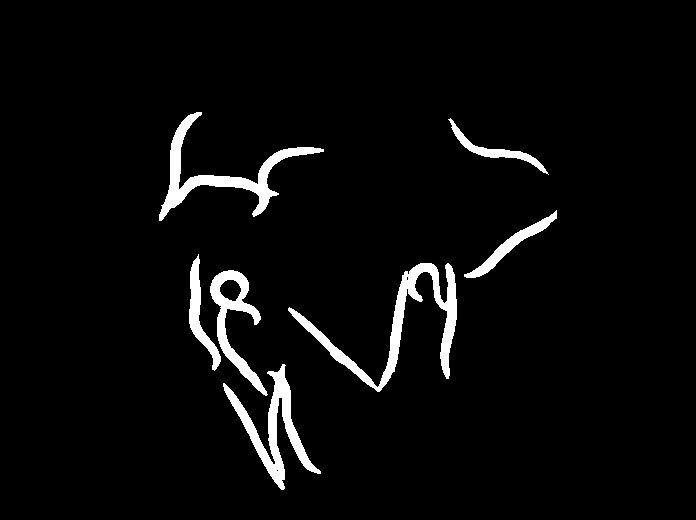}%
\includegraphics[width=0.12\textwidth,clip,trim=3cm 0 3cm 3cm]{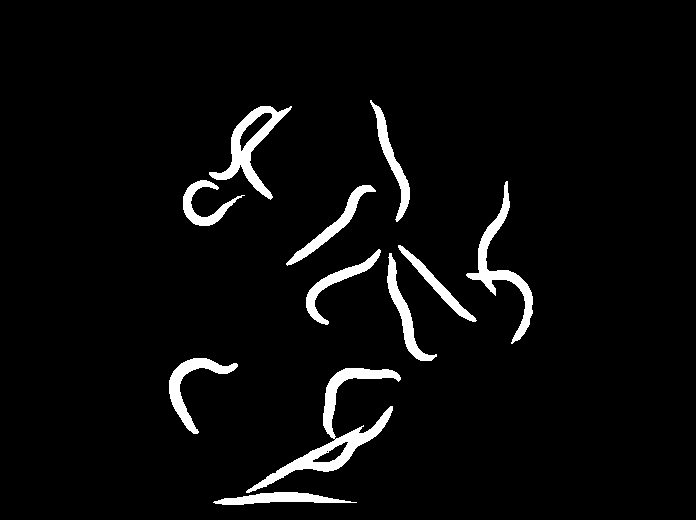}%
\includegraphics[width=0.12\textwidth,clip,trim=3cm 0 3cm 3cm]{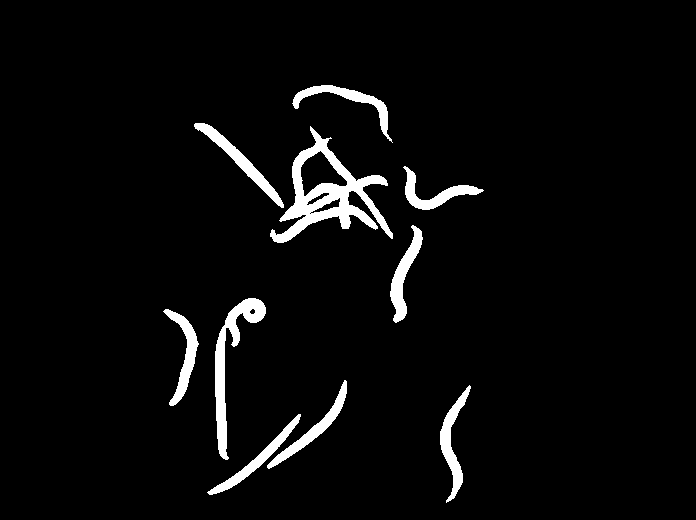}%
\includegraphics[width=0.12\textwidth,clip,trim=3cm 0 3cm 3cm]{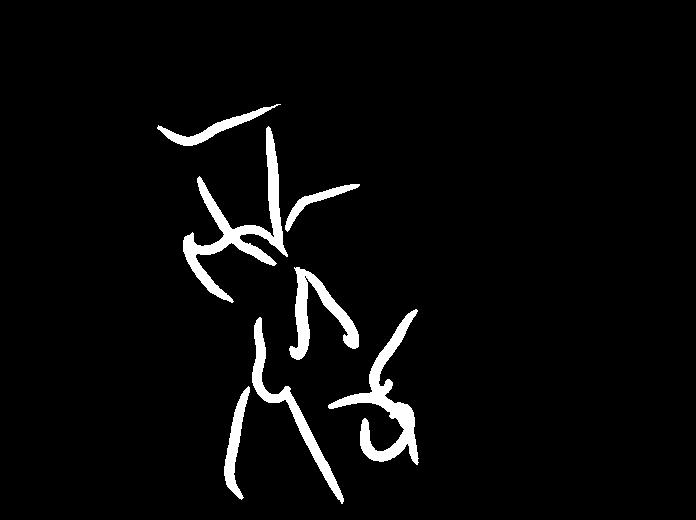}%
\includegraphics[width=0.12\textwidth,clip,trim=3cm 0 3cm 3cm]{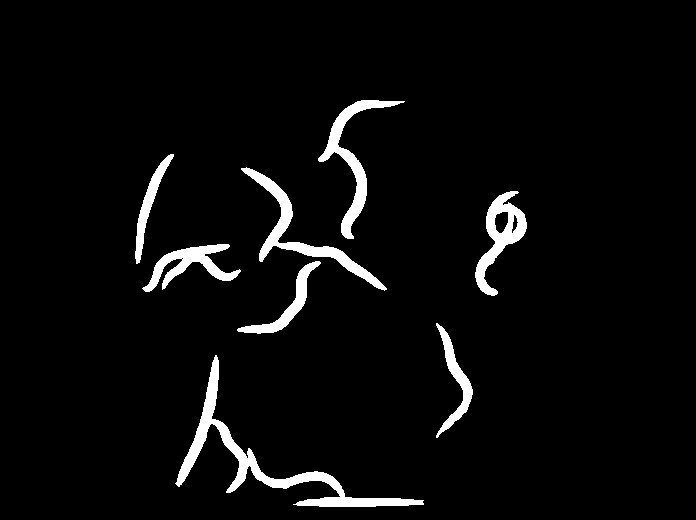}%
\\
\includegraphics[width=0.12\textwidth,clip,trim=3cm 0 3cm 3cm]{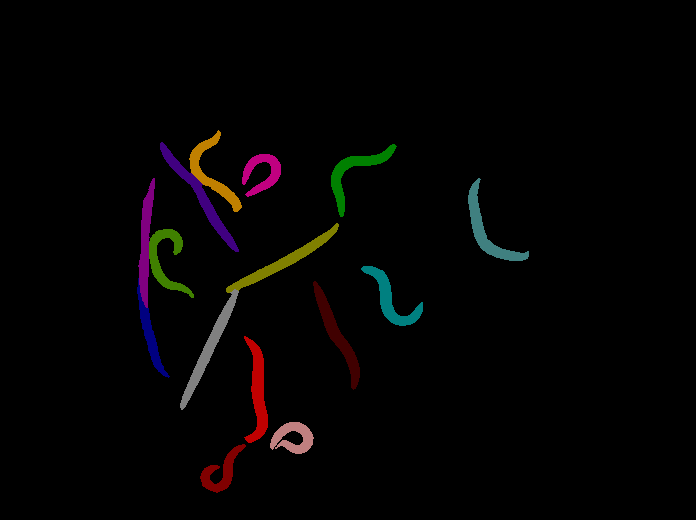}%
\includegraphics[width=0.12\textwidth,clip,trim=3cm 0 3cm 3cm]{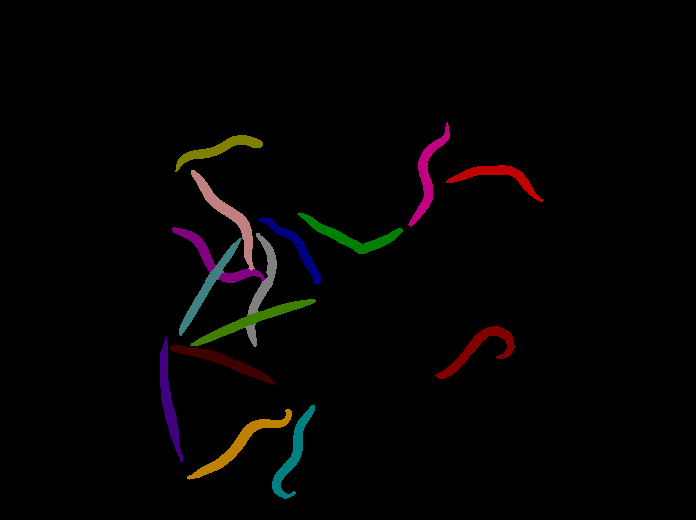}%
\includegraphics[width=0.12\textwidth,clip,trim=3cm 0 3cm 3cm]{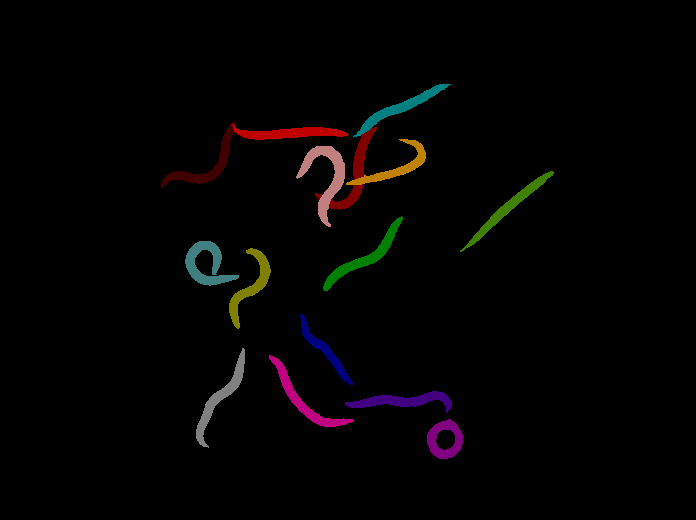}%
\includegraphics[width=0.12\textwidth,clip,trim=3cm 0 3cm 3cm]{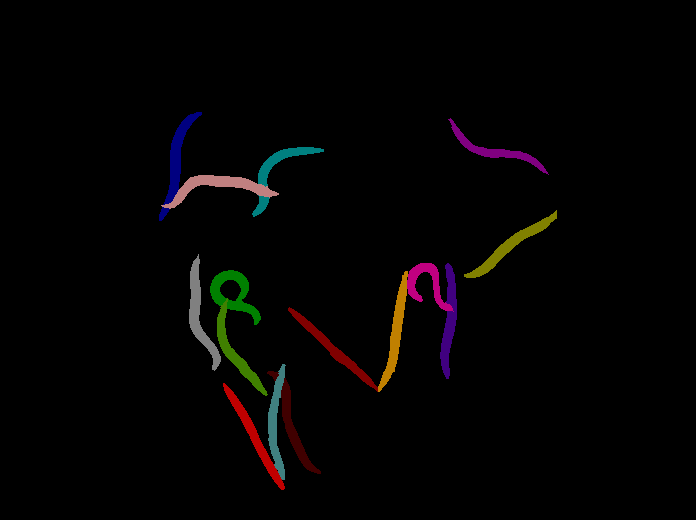}%
\includegraphics[width=0.12\textwidth,clip,trim=3cm 0 3cm 3cm]{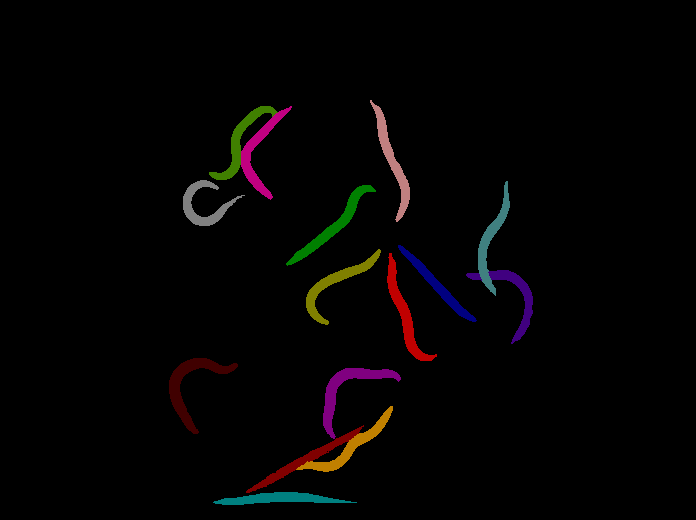}%
\includegraphics[width=0.12\textwidth,clip,trim=3cm 0 3cm 3cm]{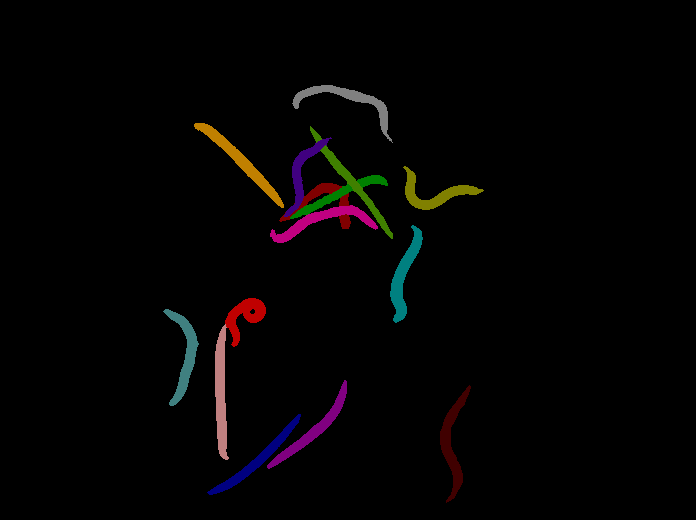}%
\includegraphics[width=0.12\textwidth,clip,trim=3cm 0 3cm 3cm]{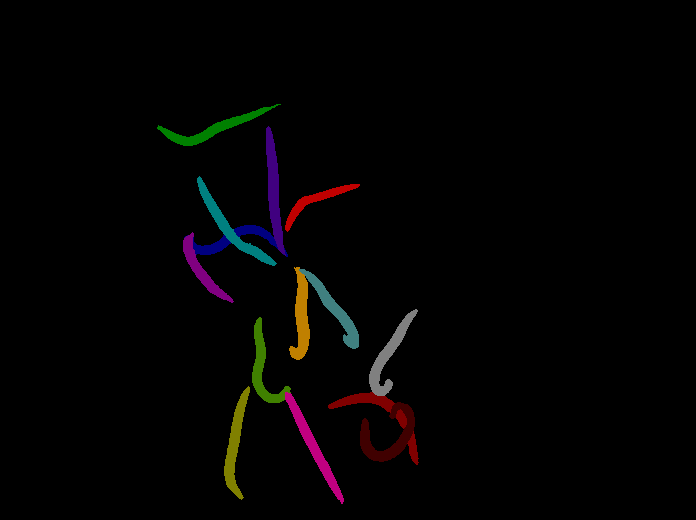}%
\includegraphics[width=0.12\textwidth,clip,trim=3cm 0 3cm 3cm]{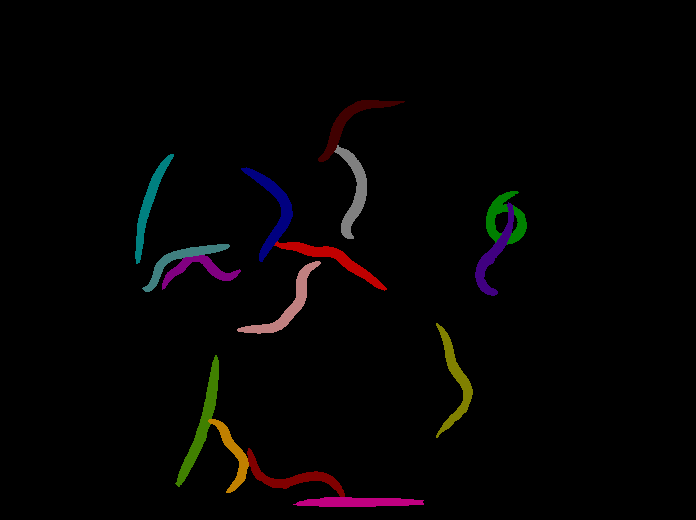}%
\end{center}
\caption{Sample image crops (top) and corresponding ground-truth (bottom) from the \textit{C. Elegans} dataset.}\label{f:elegans}
\end{figure}

\subsection{Synthetic experiments}
\label{s:synthetic}

In \cref{s:theo,s:scn} we suggested that convolution operators are unsuitable for instance segmentation via coloring, but that semi-convolutional ones can do. These experiments illustrate this point by learning a deep neural network to segment a synthetic image $x_{S}$ where object instances correspond to identical dots arranged in a regular grid (\cref{f:dots} (a)). 

We use a network consisting of a pretrained ResNet50 model truncated after the Res2c layer, followed by a set of 1$\times$1 filters that, for each pixel $u$, produce 8-dimensional pixel embeddings $\Phi_u(x_S)$ or $\Psi_u(x_S)$.  We optimize the network by minimizing the loss from $\cref{e:loss1}$ with stochastic gradient descent. Then, the embeddings corresponding to the foreground  regions are extracted and clustered with the $k$-means algorithm into $K$  clusters, where $K$ is the true number of dots present in the synthetic image.

\Cref{f:dots} visualizes the results. Clustering the features consisting of the position invariant convolutional embedding $\Phi_u(x_{S})$ results in nearly random clusters (\cref{f:dots} (c)). On the contrary, the semi-convolutional embedding $\Psi_u(x_S)=\Phi_u(x_{S})+u$ allows to separate the different instances almost perfectly when compared to the ground truth segmentation masks (\cref{f:dots} (d)).

{
\setlength{\tabcolsep}{1em} 
\begin{table}[t]
\centering
\caption{\textbf{Average precision (AP) for instance segmentation on \textit{C. Elegans}}
reporting the standard COCO evaluation metrics \cite{lin2014microsoft} }

\begin{tabular}{llllll}
  \hline
AP                        & AP    & $\text{AP}_{0.5}$ & $\text{AP}_{0.75}$ & $\text{AP}_S$ & $\text{AP}_M$ \\ \hline
Ours                      & \textbf{0.569} &\textbf{0.885} &\textbf{0.661} &\textbf{0.511} &\textbf{0.671} \\
MRCNN \cite{hu17fastmask} & 0.559 & 0.865 & 0.641 & 0.502 & 0.650 \\
  \hline
\end{tabular}
\label{t:elegans_table} 
\label{tab:elegans_table}
\end{table} 
}

\begin{figure}[t]
\begin{center}
\includegraphics[height=1.8cm]{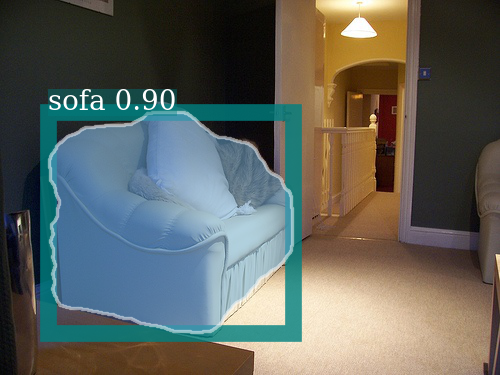}
\includegraphics[height=1.8cm]{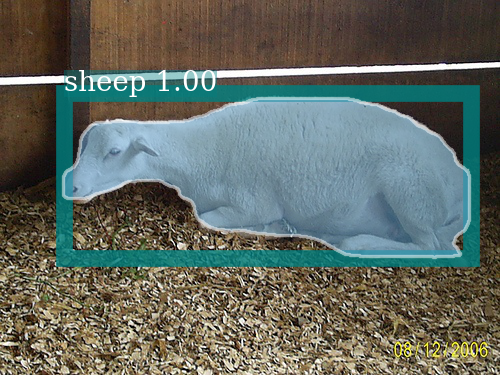}
\includegraphics[height=1.8cm]{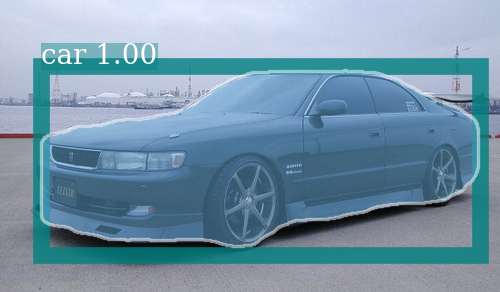}
\includegraphics[height=1.8cm]{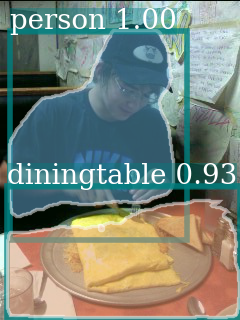}
\includegraphics[height=1.8cm]{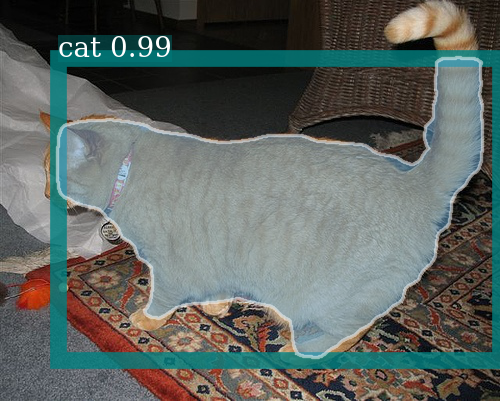}\\
\includegraphics[height=1.8cm]{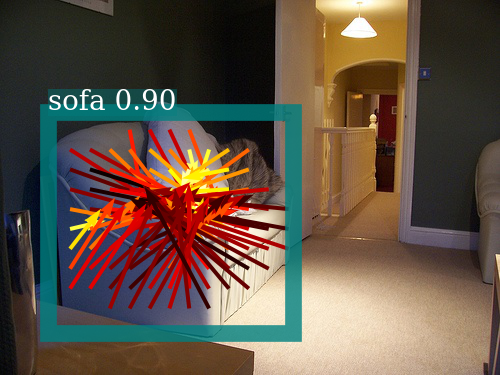}
\includegraphics[height=1.8cm]{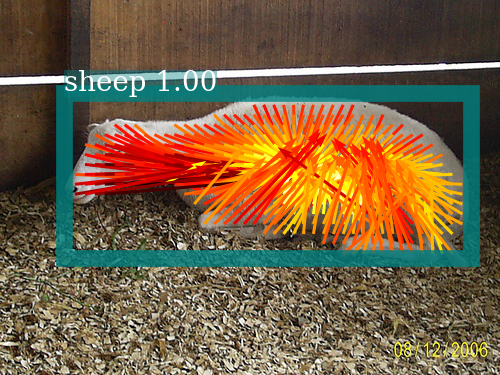}
\includegraphics[height=1.8cm]{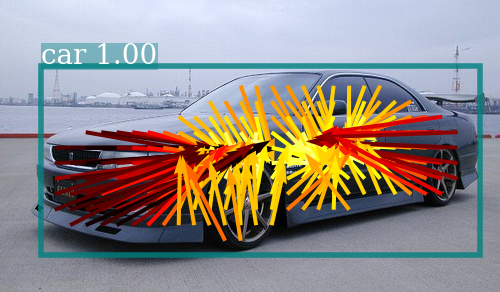}
\includegraphics[height=1.8cm]{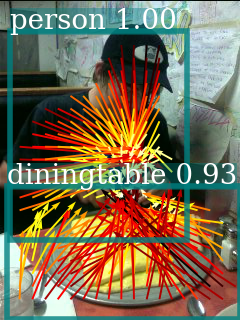}
\includegraphics[height=1.8cm]{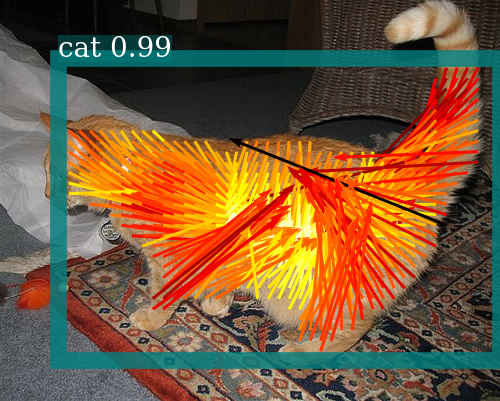}\\
\includegraphics[height=1.86cm]{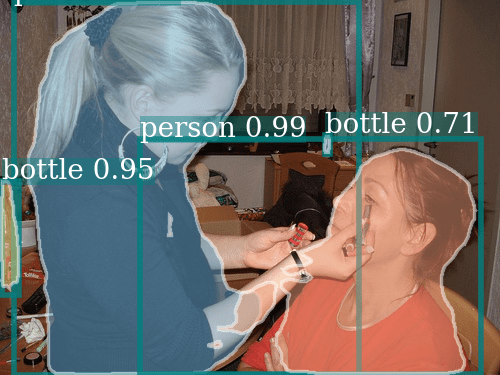}
\includegraphics[height=1.86cm]{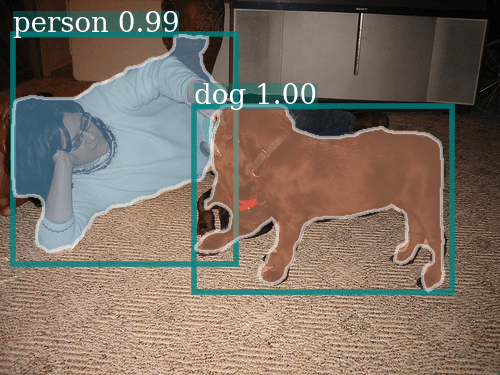}
\includegraphics[height=1.86cm]{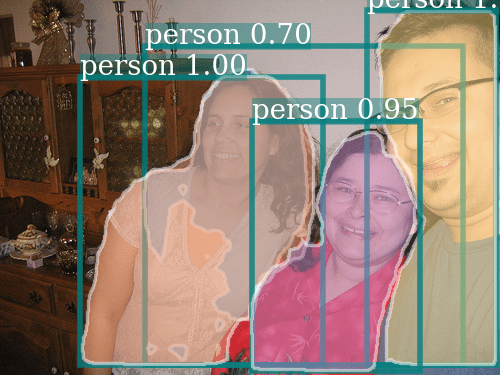}
\includegraphics[height=1.86cm]{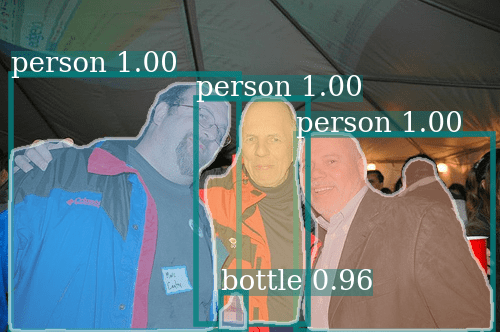}
\includegraphics[height=1.86cm]{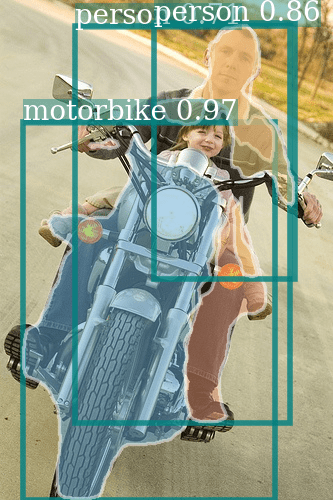}\\
\includegraphics[height=1.86cm]{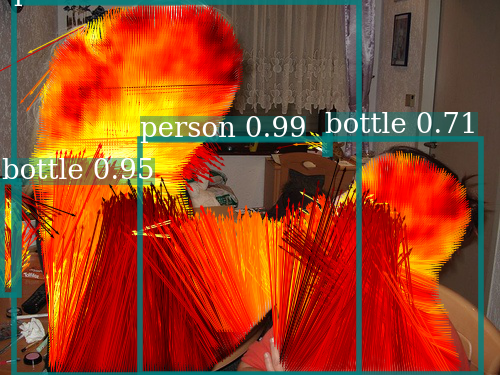}
\includegraphics[height=1.86cm]{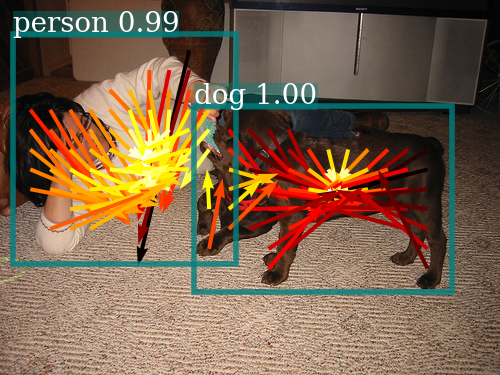}
\includegraphics[height=1.86cm]{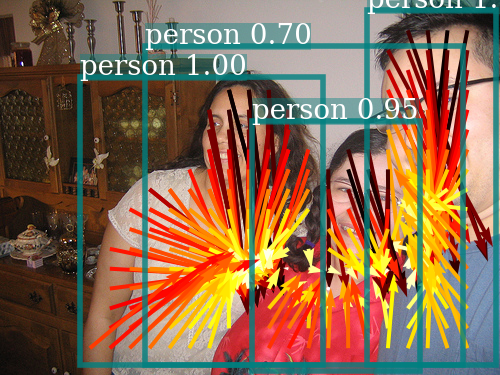}
\includegraphics[height=1.86cm]{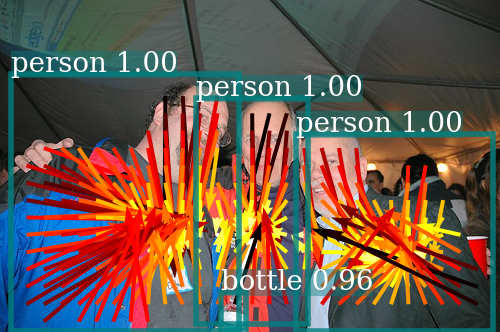}
\includegraphics[height=1.86cm]{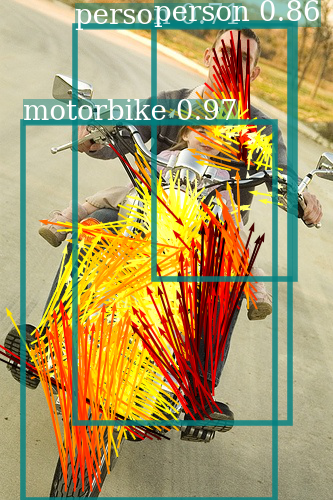}\\
\end{center}
\caption{ \textbf{Instance segmentation on Pascal VOC 2012.} Each pair of rows
visualizes instance segmentations produced with method, 
together with the corresponding semi-convolutional embeddings} 
\label{f:detections}
\end{figure}

\subsection{Parsing biological images}\label{s:biological}

The second set of experiments considers the parsing of biological images. Organisms to be segmented present non-rigid pose variations, and frequently form clusters of overlapping instances, making the parsing of such images challenging. Yet, this scenario is of crucial importance for many biological studies. 

\myparagraph{Dataset and evaluation}
We evaluate our approach on the C. Elegans dataset (illustrated \cref{f:elegans}), a subset of the Broad Biomedical Benchmark collection \cite{ljosa12annotated}. The dataset consists of $100$ bright-field microscopy images.  Following standard practice \cite{wahlby10resolving,yurchenko16parsing}, we operate on the binary segmentation of the microscopy images. However, since there is no publicly defined evaluation protocol for this dataset, a fair numerical comparison with previously published experiments is infeasible.  We therefore compare our method against a very strong baseline (MRCNN) and adopt the methodology introduced by \cite{yurchenko16parsing} in which the dataset is divided into $50$ training and $50$ test images. We evaluate the segmentation using average precision (AP) computed using the standard COCO evaluation criteria \cite{lin2014microsoft}.
We compare our method against the MRCNN FPN-101 model from \cite{he17mask} which attains results on par with state of the art on the challenging COCO instance segmentation task. 

\myparagraph{Results}
The results are given in \cref{tab:elegans_table}. 
We observe that the semi-convolutional embedding $\Psi_{u}$
brings improvements in all considered instance segmentation metrics.
The improvement is more significant at higher IoU thresholds which underlines the importance
of utilizing position sensitive embedding in order to precisely delineate an instance
within an MRCNN crop.

\subsection{Instance segmentation} \label{s:ic_pascal}
{
\begin{table}[t]
\caption{\textbf{Instance-level segmentation comparison using mean APr metric} at $0.5$ IoU on the PASCAL VOC 2012 validation set}

\begin{center}
\begin{tabular}{ l c c c c c c c c c c c c c c c c c c c c c }
  \hline
\methodsize{SDS~\cite{hariharan14simultaneous}} &
\methodsize{PFN~\cite{liang15proposal}}  &
\methodsize{DIN~\cite{arnab17pixelwise}} &
\methodsize{MNC~\cite{dai16instance-aware}} &
\methodsize{FCIS~\cite{li17fully}}  &
\methodsize{R2-IOS~\cite{liang16reversible}}  &
\methodsize{DML~{\cite{fathi17semantic}}} &
\methodsize{R. Emb.~\cite{kong18recurrent}} &
\methodsize{BAIS \cite{hayder17boundary}} &
\methodsize{MRCNN \cite{hu17fastmask}} &
\methodsize{\textbf{Ours}} \\ \hline
43.8 & 58.7 & 61.7 & 63.5 & 65.7 & 66.7 & 62.1 & 64.5 & 65.7 & 69.0 & \textbf{69.9}\\
  \hline
\end{tabular}
\end{center} 
\label{tab:pascal_table}
\end{table} 
}

{
\setlength{\tabcolsep}{1em} 
\begin{table}[t]
\centering
\caption{\textbf{Average precision (AP) for instance segmentation on PASCAL VOC 2012}
reporting the standard COCO evaluation metrics \cite{lin2014microsoft}}
\begin{tabular}{lrrrrrr}
  \hline
AP                        & $AP$    & $\text{AP}_{0.5}$ & $\text{AP}_{0.75}$ & $\text{AP}_S$ & $\text{AP}_M$ & $\text{AP}_L$ \\ \hline
Ours                      & \textbf{0.412} & \textbf{0.699} & \textbf{0.424} & 0.107 & \textbf{0.317} & \textbf{0.538}  \\
MRCNN \cite{hu17fastmask} & 0.401 & 0.690 & 0.412 & \textbf{0.111} & 0.313 & 0.525 \\
\hline
\end{tabular}
\label{tab:pascal_ablation}
\end{table}
}
 
The final experiment compares our method to competition on the instance segmentation task on a standard large scale dataset, PASCAL VOC 2012 \cite{pascal-voc-2012}.

As in the previous section, we base our method on the MRCNN FPN-101 model. Because we observed that the RPN component is extremely sensitive to changes in the base architecture, we employed a multistage training strategy. First, MRCNN FPN-101 model is trained until convergence and then our embeddings are attached and fine-tuned with the rest of the network . We follow~\cite{he17mask} and learn using 24 SGD epochs, lowering the initial learning rate of 0.0025 tenfold after the first 12 epochs. Following other approaches, we train on the training set of VOC 2012 and test on the validation set.

\myparagraph{Results}
The results are given in \cref{tab:pascal_table}. Our method attains state of the art on PASCAL VOC 2012 which validates our approach. We further compare in detail against MRCNN in \cref{tab:pascal_ablation} using the standard COCO instance segmentation metrics from \cite{lin2014microsoft}. Our method outperforms MRCNN on the considered metrics, confirming the contribution of the proposed semi-convolutional embedding.

\section{Conclusions}\label{s:conc}

In this paper, we have considered dense pixel embeddings for the task of instance-level segmentation. Departing from standard approaches that rely on translation invariant convolutional neural networks, we have proposed semi-convolutional operators which can be easily obtained with simple modifications of the convolutional ones. On top of their theoretical advantages, we have shown empirically that they are much more suited to distinguish several identical instances of the same object, and are complementary to the standard Mask-RCNN approach.

\myparagraph{Acknowledgments}
We gratefully acknowledge the support of Naver, EPSRC AIMS CDT, AWS ML Research Award, and ERC 677195-IDIU.

\bibliographystyle{splncs04}
\bibliography{refs}

\end{document}